\documentclass[a4size,conference]{IEEEtran}
\usepackage{amsmath,amsfonts}
\usepackage{algorithmic}
\usepackage{algorithm}
\usepackage{array}
\usepackage[caption=false,font=normalsize,labelfont=sf,textfont=sf]{subfig}
\usepackage{textcomp}
\usepackage{stfloats}
\usepackage{booktabs}
\usepackage{url}
\usepackage{verbatim}
\usepackage{graphicx}
\usepackage{cite}
\begin{document}

\title{Next Best View For Point-Cloud Model Acquisition: Bayesian Approximation and Uncertainty Analysis}

\author{\IEEEauthorblockN{Madalena Pombinho Caeiro Caldeira}
\IEEEauthorblockA{Instituto Superior Técnico\\
University of Lisbon\\
%Atlanta, Georgia 30332--0250\\
Email: madalenacaldeira@tecnico.ulisboa.pt}
\and
\IEEEauthorblockN{Plinio Moreno}
\IEEEauthorblockA{Instituto Superior Técnico\\
Institute for Systems and Robotics, LARSyS\\
University of Lisbon\\
Email: plinio@isr.tecnico.ulisboa.pt}}
\maketitle

\begin{abstract}
The Next Best View problem is a computer vision problem widely studied in robotics. To solve it, several methodologies have been proposed over the years. Some, more recently, propose the use of deep learning models. Predictions obtained with the help of deep learning models naturally have some uncertainty associated with them. Despite this, the standard models do not allow for their quantification. However, Bayesian estimation theory contributed to the demonstration that dropout layers allow to estimate prediction uncertainty in neural networks.

This work adapts the point-net-based neural network for Next-Best-View (PC-NBV). It incorporates dropout layers into the model's architecture, thus allowing the computation of the uncertainty estimate associated with its predictions.
The aim of the work is to improve the network's accuracy in correctly predicting the next best viewpoint, proposing a way to make the 3D reconstruction process more efficient.

Two uncertainty measurements capable of reflecting the prediction's error and accuracy, respectively, were obtained. These enabled the reduction of the model's error and the increase in its accuracy from 30\% to 80\% by identifying and disregarding predictions with high values of uncertainty. Another method that directly uses these uncertainty metrics to improve the final prediction was also proposed. However, it showed very residual improvements.
\end{abstract}


\section{Introduction}
\IEEEPARstart{D}{evelopments} in the fields of robotics and artificial intelligence have been allowing several improvements in some scientific and non-scientific areas, revolutionizing everyday tasks in the process. This influence can also be visible in the domains of archaeology, conservation and restoration, where their strategic integration has become an invaluable asset, aiding men in tasks that require lots of precision and time investment. The RePair project \cite{repair_project} is proof of this. It focuses on the use of intelligent robotics in this area, with the goal of making the reconstruction process of fragmented Pompeii frescoes more efficient. Some methodologies for reconstruction rely on computational help - it is possible to use computer algorithms that, just like solving a jigsaw puzzle, are able to perform the reconstructions. For this, however, they require the 3D models of the fragments, which is the task that inspired this work. We study an algorithm for the efficient 3D model acquisition of the fresco fragments.

The 3D data collection method must be as efficient as possible. This means that the algorithm needs to find a way of getting the most information of the fragments surface with as few scans as possible.
To achieve this goal the algorithm has to find the camera viewpoints (i.e. camera pose with respect to the object being scanned) that will get the most new information on the object being scanned. This is known as the Next Best View (NBV) problem, which the algorithm must adeptly solve to achieve its intended outcome. Recent approaches are based on Deep Learning architectures such as \cite{NBV-net}, \cite{NBV-net_Regr}, \cite{PC-NBV} and \cite{DB_NBV}, showed very good results in performing 3D model reconstructions in an efficient manner. However, they do not account for the uncertainty of their predictions which, in real world scenarios, can be vital to guarantee a good reconstruction performance.% After all, a prediction with high uncertainty will most likely be a wrong prediction that can potentially jeopardize the reconstruction efficiency.

Hence, in this work, a NBV that predicts the coverage of the viewpoints - PC-NBV \cite{PC-NBV} - is analyzed in a more detailed way, and modifications are made to its architecture to make it account for uncertainty.
The uncertainty of the estimated coverage scores is calculated using the idea of Monte Carlo sampling to estimate the uncertainty of the output of a Neural Network. To do this, the deterministic output becomes probabilistic through the application of dropout layers. 
The goal is for the model to output a prediction along with an uncertainty measurement that reflects how sure the model is of its predictions, in an attempt of enhancing precision with the information given by this measurement. This can subsequently improve the efficiency of the 3D model acquisition process.

Two types of metrics that give two different kinds of uncertainty measurements were obtained. One is error related, as it estimates the prediction uncertainty of the coverage estimation. Another one is accuracy related, as it estimates the prediction uncertainty related to the ability of the network on choosing the right NBV. These metrics allow to improve the performance of the NBV selection.

\section{State of the Art} 
\subsection{Deep Learning for the NBV Problem Resolution}
The vast majority of deep learning-based methods for NBV, usually approximate the continuous space by creating a discrete set of possible camera viewpoints, the candidate viewpoints. Each method then has its own set of evaluation criteria to measure the quality of each candidate viewpoint. Deep learning enters this process by learning how to approximate the function that measures their quality, being then able to predict it. Amongst previous works \cite{NBV-net}, \cite{NBV-net_Regr}, \cite{PC-NBV} and \cite{DB_NBV}, we selected the Point Cloud Based Next-Best-View Network (PC-NBV) \cite{PC-NBV} that relies on point cloud for the object representation \cite{pointnet,pointnet++}. Point cloud representation is less computationally costly compared to other common representation methods like voxels and triangular meshes.
%This network uses deep learning not only to predict the NBV but also for point cloud processing, as deep learning has proven to be quite successful in this area \cite{pointnet}, \cite{pointnet++}.

The network receives a raw point cloud and selects what is the NBV out of a set of 33 candidate views. The candidate views are defined by uniformly sampling points on a sphere around the object. On each view, the coverage score (that measures the amount of new information about the object each view can capture) is computed as:
\begin{equation}
    C(P)=\frac{1}{|P_o|}\sum_{p\in P}^{}U(min_{p_0\in P_0}||p-p_0||_2+\varepsilon),
    \label{eq:coverage_score}
\end{equation}
where $U$ is the Heaviside step function, $\epsilon$ is a distance threshold, $P_0$ is the complete point cloud of the object and $P$ is the object's partial point cloud - the point cloud of the object's reconstruction. Essentially this score quantifies the amount of new points a viewpoint captures in relation to the object partial point cloud (points of the object's total point cloud that have not yet been included in the partial point cloud). It presents the value as a ratio between the number of new points and the total number of points of the object's full point cloud.
 
\begin{figure}[H]
     \centering
     \includegraphics[width=\columnwidth]{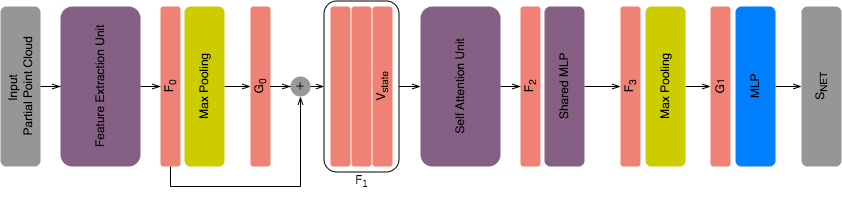}
     \caption{PC-NBV Units Architecture. Convolutional Layers: Orange; Activation Layers: Green; Inputs: Gray; Features: Pink}
     \label{fig:PC_NBV_network}
 \end{figure}

 The network's architecture is represented in the Figure \ref{fig:PC_NBV_network}. It consists firstly of a feature extraction unit proposed in \cite{patch-based}, used to extract structure-aware features. %\cite{patch-based}'s ablation study showed that this unit improved the network's capacity, contributed for a significant reduction of the number of parameters, and showed that it extracted better features from the point cloud.  
 The unit is then followed by a max pooling layer and a self attention unit proposed in \cite{self-att} %According to this study, attention mechanisms have become an integral part of models that have to capture global dependencies.
 that enables the network to learn how to allocate attention according with similarity of color and texture. This unit is then followed by a shared multilayer perceptron, a max pooling layer and another multilayer perceptron. 

 The total loss of the model is given by
 \begin{equation}
\mathcal{L} = \frac{1}{n_s} \sum_{i=1}^{n_s} (y_i - \hat{y}_i)^2 + \lambda \sum_{j=1}^{M} w_j^2,
\label{eq:loss}
\end{equation}
which is the sum of the MSE between the coverage score prediction ($\hat{y}$) and the groundtruth ($y$) with the L2 loss of the weights ($w_j$). The MSE between the coverage score prediction and the groundtruth measures how accurate the model is. The L2 regularization loss measures how well the model adheres to the regularization constraints. This network shows great results compared with other NBV methods, both in accuracy, inference time and generalization to unseen objects \cite{PC-NBV}. It quickly reaches high surface coverage levels both with objects similar to the ones analyzed in the training data, as with objects that the network has never seen before. Also, the inference time of PC-NBV is smaller than in other methods, because the point cloud representation does not require any complex data transformations like ray projection operations that other methods require. 

\subsection{Uncertainty in Deep Learning}
%Deep learning models are not always certain about their predictions. Taking for example a model that is fed with a test point that is out of the distribution of the set it was trained on: it is forced to output a prediction for this point. However, it does not have enough training data to support the decision. The probabilities of it resulting in a wrong prediction are high, the model makes a prediction with high uncertainty. Hence, in some applications, it can be important to know the model's prediction with an extra value that measures how certain the model is of its output.

The standard tools used to create deep learning models, like feedforward neural networks and convolutional neural networks, cannot usually access the model's confidence on its predictions. 
Bayesian networks \cite{bayesian_networks} allow the computation of the predictions confidence by considering the weights of the networks as probabilistic distributions. This offers a mathematically grounded framework to reason about model uncertainty. In these networks, in each forward pass the network assigns a different value for the weights, sampled from the weight's distribution. In practice, in each forward pass there is a different learned function $f$. Hence, with the same data a different prediction $y'=f(x,\theta)$ is made. If several forward passes are made with the same data, there are several different predictions as well, and in the end the network's final prediction can be given as an average of all the predictions, as follows:
\begin{equation}
    y'= \frac{1}{N}\sum_{i=1}^{N}f_i(x,\theta_i),
    %\frac{f_1(x,\theta_1)+f_2(x,\theta_2)+...+f_n(x,\theta_n)}{n}
    \label{eq:y_pred}
\end{equation}
and the uncertainty is obtained from the variance of $y'$.%Having several predictions it becomes possible to measure the prediction's uncertainty, simply by calculating the variance.

\cite{dropout_baye_appr} states that dropout layers can emulate the Bayesian Network's behavior. In a network with dropout layers each neuron has some $p$ probability (dropout rate) of being switched off. It is as if we get a slightly different network each time, which means we can get several different predictions as well. Most commonly, dropout layers are only used during training since they are being used to reduce overfitting \cite{dropout_layer}. However, in this method, dropout layers also have to be used during inference. This way we get several predictions to average and are able to calculate the variance. This method is referred to as the Monte Carlo Dropout Method (MCDM). This idea has been applied to segmentation \cite{bayesian_seg-net} and localization \cite{bayesian_pose-net}, amongst others.

%Many articles such as \cite{bayesian_seg-net} and \cite{bayesian_pose-net} have already successfully used the MCDM to modify already existing deep learning networks to make them account for uncertainty. Both works achieved positive results. In \cite{bayesian_seg-net} modeling uncertainty improved segmentation performance by 2-3\%, and in \cite{bayesian_pose-net} it enabled an improvement in localization accuracy. 

\section{Methodology}
To address the NBV deep learning models limitation of not being able to quantify the prediction uncertainty, and to improve the performance of the models, this work proposes modifying PC-NBV as presented in \cite{dropout_baye_appr} - making it account for uncertainty. The modification effects on the prediction will be examined, and two uncertainty-based metrics are going to be studied to improve view selection. The uncertainty measurements will be analyzed by evaluating their relevance as decision-support measures, and analyzing their potential ability to improve the model's final coverage score prediction.

\begin{figure*}[h!]
    \centering
    \includegraphics[width=\textwidth]{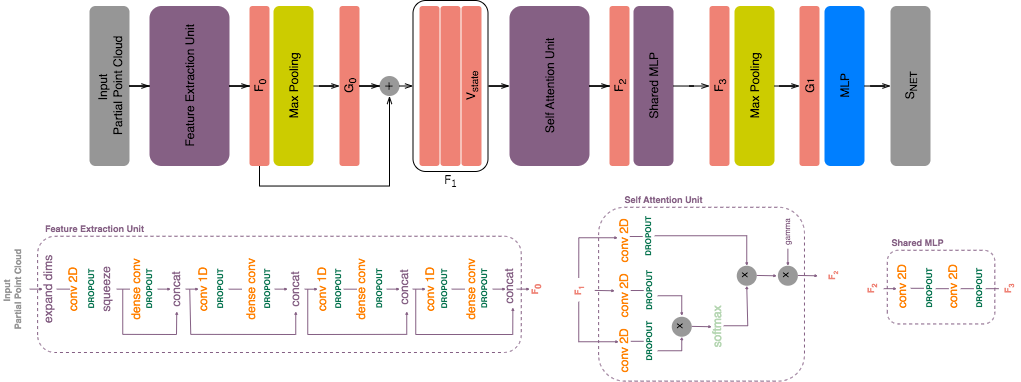}
    \caption{Bayesian PC-NBV architecture. Input and Output: Gray; Larger Units: Purple; Pooling Layers: Yellow; Fully Connected Layers: Blue; Features: Pink; Convolutional Layers: Orange; Activation Layers: Light Green; Dropout Layers: Dark Green.}
    \label{fig:bay_pc_arch}
\end{figure*}

\subsection{Dataset}
The PC-NBV's authors did not provide the partitions of the training/testing set they used to train the model, but made available the process they followed to generate it. Therefore, we followed the same process to create the training and testing datasets, from ShapeNet \cite{shapenet}, following the simulated NBV search process described in \cite{PC-NBV}.

%The adopted approach to generate the dataset was by making a simulated 3D reconstruction of objects whose full 3D models are previously known. These models were obtained from ShapeNet \cite{shapenet} - a dataset with hundreds of 3D models organized in several categories (called classes) of objects. 

For the training set 4000 models from 8 categories (airplane, cabinet, car, chair, lamp, sofa, table and vassel) were randomly chosen (500 models from each category), and another different 400 models for the validation set (50 models from each category).
Regarding inference two sets were created - one with 400 models selected from these 8 categories known in the training/validation sets, and another set - the novel set - with 400 models from other 8 categories not learned by the model. This last testing set was used to evaluate the performance of the model on unknown objects.

\subsection{Training and Model Architecture}
\subsubsection{Baseline}
To be able to evaluate the results of the network that is going to be created by modification of the PC-NBV model, it is first important to get a baseline to compare the results to. This baseline is given by the results of the original model. So it is important to train and test the original model on the created dataset.

\subsubsection{Architectural Changes}

To enable the model to account for uncertainty, by implementing the MCDM, dropout layers need to be applied after each convolutional layer. The new model architecture is presented in Figure \ref{fig:bay_pc_arch}, where the added dropout layers are signalized in dark green. This new architecture will be referred to as Bayesian PC-NBV.

% \begin{figure}
%     \centering
%     \includegraphics[width=\columnwidth]{Images/new_bayesian PC-NBV.png}
%     \caption{Bayesian PC-NBV architecture. Input and Output: Gray; Larger Units: Purple; Pooling Layers: Yellow; Fully Connected Layers: Blue; Features: Pink; Convolutional Layers: Orange; Activation Layers: Light Green; Dropout Layers: Dark Green.}
%     \label{fig:bay_pc_arch}
% \end{figure}

When setting the dropout layers, a dropout probability needs to be defined. Usually, the dropout probability falls between $0.2$ and $0.5$. We follow previous studies that verified empirically that $0.5$ is a good value \cite{bayesian_pose-net, bayesian_seg-net}.

\subsubsection{Performance evaluation metrics}
In this dissertation the performance of the models is evaluated with two metrics besides the loss \eqref{eq:loss}. These two measurements are the error and accuracy of the predictions. The error measures the difference between the model prediction and the groundtruth. It differs from the loss since it does not account for the regularization loss. The accuracy measures if the model's predicted NBV matches the groundtruth's NBV.

Various error metrics were evaluated, in order to pick the one that would be the most relevant.
\begin{equation}
    Euclidean\:Distance = \lvert gt - fp \lvert = \sqrt{\sum_{i=1}^{n_v} \left(gt_{i} - fp_{i}\right)^2}
    \label{eq:dist_error}
\end{equation}
measures the error by calculating the Euclidean Distance between the groundtruth ($gt$) and the model prediction ($fp$). 
\begin{equation}
    Squared\:Error = \sum_{i=1}^{n_v} (gt_{i} - fp_{i})^2 = Euclidean\:Distance ^ 2
    \label{eq:mse_error}
\end{equation}
measures the sample error by calculating the sum of each view Squared Difference between $gt$ and $fp$, which is the square of the Euclidean Distance. The model error is given by the mean of the sample errors.

%Despite the fact that PC-NBV is a regression model, the scores it regresses are used to choose a NBV, i.e - in a real world problem the closeness of the coverage score predictions to the groundtruth is not highly important, the most important aspect is that the model predicts the highest coverage score on the right viewpoint. And for this is important to evaluate if the model has the ability to predict the right NBV viewpoint or not.
Although the PC-NBV architecture is a regression model, the view selection performs a maximum-pooling like operation to select the view. Thus, it is important to evaluate the accuracy of the view selection. Furthermore, in some cases the maximum coverage value is close between several views, so there are several views with similar coverage gain. To include these cases, we defined an interval given by \eqref{eq:intervalo}, where all those similar coverage values are considered as correct ones. Then, the accuracy computes the ratio between the number of right predictions and of total samples \eqref{eq:model_accuracy}:
%In most of the cases, in the groundtruth, there is only one NBV - the viewpoint with the highest coverage score. However, up on further examination, when visualizing the data, it is possible to understand that in some cases more than one view achieve coverage scores close to the NBV coverage score value.
%To calculate the accuracy of each sample this circumstance is taken into consideration. We analyze if the predicted NBV belongs to a set of views whose coverage scores lay in an interval given by \eqref{eq:intervalo}, i.e. a set of top views with close coverage scores between each other. If the model chooses a viewpoint that is inside this set we assume the prediction is right, if not the prediction is set as wrong -  \eqref{eq:sample_accuracy}. In the end the model accuracy is given by the ratio between the number of right predictions and the total number of samples - \eqref{eq:model_accuracy}.
%The graph in Figure \ref{fig:NBV_maximums} allows for a better comprehension of what is considered a right prediction - the viewpoints that lay on the gray area are the ones considered as NBVs.

\begin{equation}
    accuracy(fp)= \begin{cases}
        1, \text{if $fp_{NBV}$ $\in$ \{NBVS}\} \\
        0, \text{otherwise}
    \end{cases}
    \label{eq:sample_accuracy}
\end{equation}
\begin{equation*}
    \{NBVS\} = \{v_{i^*}\}, \{i^*\} = \{i^* \in arg(C),
\end{equation*} 
\begin{equation}
    C \in [max(fp) - 0.15\alpha, max(fp)]\}
    \label{eq:intervalo}
\end{equation}
\begin{equation*}
    \alpha = max(C) - min(C)
\end{equation*}

\begin{equation}
    model\:accuracy= \frac{\sum_{i=1}^{n_s} accuracy(fp_i)}{n_s}
    \label{eq:model_accuracy}
\end{equation}

%\begin{figure}
%    \centering
%    \includegraphics[width=3in]{Images/new_NBVS_3.png}
%    \caption{Coverage Socre groundtruth plot with signalized NBVS interval.}
%    \label{fig:NBV_maximums}
%\end{figure}

\subsection{Bayesian PC-NBV Inference and Uncertainty}
\subsubsection{Inference}
The usual behavior of dropout layers is to be on during testing and off during inference. However, in this case, to get several different predictions for the same test sample, it is desired that the dropout layers are on during inference as well.

The MCDM requires a somewhat large number of inferences of the same test sample. Based on the conclusions of the works \cite{bayesian_seg-net, bayesian_pose-net} this number ($n_{mc}$) was set to 40. The 40 MC samples ($mc$) are then averaged to obtain the model's final prediction:

\begin{equation}
    fp = \hat{C(P)}= \frac{\sum_{i=1}^{n_{mc}} mc_{i}(x)}{n_{mc}}
    \label{eq:final_pred}
\end{equation}

\subsubsection{Uncertainty Measurements}
The uncertainty measurements are given by the differences along the MC samples. In theory, samples that show larger differences between themselves reflect a higher model uncertainty.

The model's prediction is composed of a coverage score for each of the 33 viewpoints. So it is possible to calculate the standard deviation for each viewpoint, having an uncertainty metric for each view. However, it would also be interesting if the model could return a single value that reflects how certain the model is of its prediction. To address this question several standard deviation metrics were studied: Standard Deviation by View, $\sigma_v$ \eqref{eq:std_by_view}; Mean Standard Deviation, $\overline{\sigma}$ \eqref{eq:std_mean}; NBV Standard Deviation, $\sigma_{NBV}$ \eqref{eq:std_nbv}; Standard Deviation as a Whole, $\sigma_{whole}$ \eqref{eq:std_whole}:

\begin{equation}
\begin{split}
    \sigma_{v_{j}} = \sqrt{\frac{1}{n_{mc}}\sum_{i=1}^{n_{mc}}\left( fp(v_j) - mc_i(v_j) \right)^2}, j = \{ 1, ..., n_v\},
    %std \: by \: view (x) = \sqrt{\frac{\sum_{i=1}^{40} (fp(x) - p_{i}(x))^2}{40}}
    \label{eq:std_by_view}
\end{split}
\end{equation}
\begin{equation}
    \overline{\sigma_v} = \frac{1}{n_v}\sum_{j=1}^{n_v}\sigma_{v_{j}},
    %std\:mean(x) = \frac{1}{33} \sum_{v=1}^{33} (std\:by\:view_v(x))
    \label{eq:std_mean}
\end{equation}
\begin{equation}
    \sigma_{NBV} = \sigma_{v_{i*}}, v_{i*} = argmax(fp)
    %std\:nbv(x) = std\:by\:view_{v_{i*}}(x)
    \label{eq:std_nbv}
\end{equation}
% \begin{equation*}
%     v_{i*} = argmax(fp)
% \end{equation*}
\begin{equation}
\begin{split}
    %\sigma_{whole} = \sqrt{\frac{1}{n_{mc}} \sum_{i=1}^{n_{mc}}\lvert fp - mc_{i} \lvert ^2} = \\
    \sigma_{whole} = \sqrt{\frac{1}{n_{mc}} \sum_{i=1}^{n_{mc}} \left[ \sum_{j=1}^{n_v} \left( fp(v_j) - mc_i(v_j) \right) ^2 \right]}.
    %std\:whole(x) = \sqrt{\frac{\sum_{i=1}^{40} \lvert fp(x) - p_i(x) \lvert ^2}{40}} =  \\
    %= \sqrt{\frac{\sum_{i=1}^{40} [\sum_{v=i}^{33} (fp_v(x) - p_{i_v}(x))^2]}{40}}
    \label{eq:std_whole}
\end{split}
\end{equation}

%All of these metrics were evaluated to understand which one could better translate the uncertainty and better adapted to this problem.

\section{Results Analysis}
\subsection{Training and Testing Results}

\begin{figure}[H]
    \centering
    \includegraphics[width=2.8in]{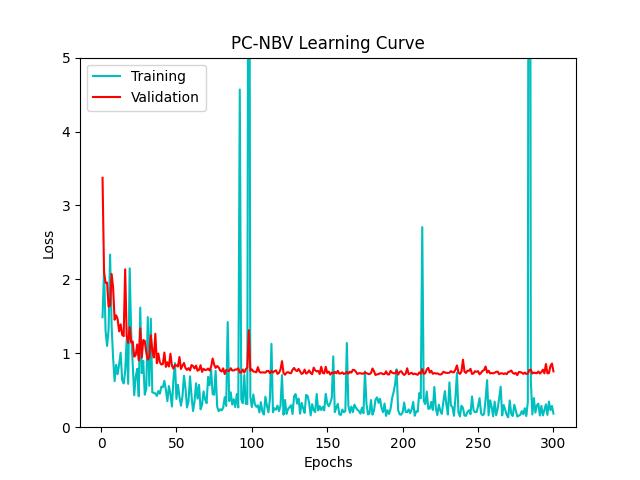}
    \caption{PC-NBV Learning Curve. Loss given by Equation \eqref{eq:loss}}
    \label{fig:pc-nbv-learn-curve}
\end{figure}

The PC-NBV model was trained in the training set, its learning curve is shown in Figure \ref{fig:pc-nbv-learn-curve}, it is visible that the model was able to learn the data without overfitting it.

The loss, error and accuracy results of both models are presented in Tables \ref{tab:pc-nbv-results} and \ref{tab:bay-pc-nbv-results}, these are the metrics that are going to be used to compare model performances.

%The Bayesian PC-NBV model was trained with the same parameters used for the PC-NBV model. The model's learning curve is presented in Figure \ref{fig:bay-pc-nbv-learn-curve}. It is visible that also here the model did not overfitt the data. However, compared to the previous training process the loss seems now slightly more unstable. This is an expected behavior caused by the randomness dropout introduces during the training process, causing the loss to fluctuate more during training compared to when dropout is not used. Table \ref{tab:bay-pc-nbv-results} shows the network's loss, error and accuracy values.

% \begin{figure}[H]
%     \centering
%     \includegraphics[width=2.8in]{Images/BAYESIAN-PC-NBV-LEARN-CURVE.png}
%     \caption{Bayesian PC-NBV Learning Curve. Loss given by Equation \eqref{eq:loss}.}
%     \label{fig:bay-pc-nbv-learn-curve}
% \end{figure}

Comparing these tables it is visible that there are very little performance shifts between the models. The loss throughout the sets keeps quite similar values, with a small increase of about 0.1 in some sets. The model error values virtually do not change between both models. When it comes to the model accuracy there is a slight decrease in the performance of the Bayesian PC-NBV model, as it was verified a decrease of about 1 percentage point across all the sets.

%\begin{figure}
%    \includegraphics[width=3in]{Images/mc-samples-3.png}
%    \caption{Plot of the Monte Carlo samples (in colors), their mean (in black) and the groundtruth (in blue).}
%    \label{fig:mc-samples-plots}
%\end{figure}

%Figure \ref{fig:mc-samples-plots} shows in colors the plot of the 40 MC samples for a test samples, in black their mean - the denominated final prediction \eqref{eq:final_pred} - and in blue the sample's ground truth. While observing this plots it is possible to conclude that the dropout was in fact active during inference - which explains the difference in model predictions for the same input data. It is also interesting to notice that the shape of the predictions is not always the same throughout the 40 inferences, meaning the model does not always predicts the same NBV, which will be relevant in the quantification of the prediction's uncertainty. 

Therefore, despite the introduction of dropout layers and the inherent randomness they entail during training, the model keeps presenting a good performance when implementing the Monte Carlo sampling method during inference.

%Although some  marginal shifts in loss, error, and accuracy metrics were verified, when comparing to the PC-NBV model, these were quite small and are not considered as a big performance drop. These small changes are the price to pay to enable the model to predict an uncertainty measurement, which will be addressed next.

\begin{table}[h]
\caption{PC-NBV Model Loss, Model Error - Euclidean Distance and Squared Differences, and Model Accuracy}
\centering
\begin{tabular}{|c|c|c|c|c|}
\hline
\textbf{PC-NBV} & Loss, \eqref{eq:loss} & Error, \eqref{eq:dist_error} & Error, \eqref{eq:mse_error} & Accuracy, \eqref{eq:model_accuracy}  \\ \hline
Train   & 0.144     & -     & -     & - \\ \hline
Valid   & 0.691     & 0.070     & 0.018     & 29.76\%        \\ \hline
Test    & 0.662     & 0.069     & 0.017     & 29.95\%        \\ \hline
Test Novel & 1.160  & 0.093     & 0.033     & 26.59\%        \\ \hline
\end{tabular}
\label{tab:pc-nbv-results}
\end{table}
\begin{figure*}[h]
    \centering
    \includegraphics[scale=0.55]{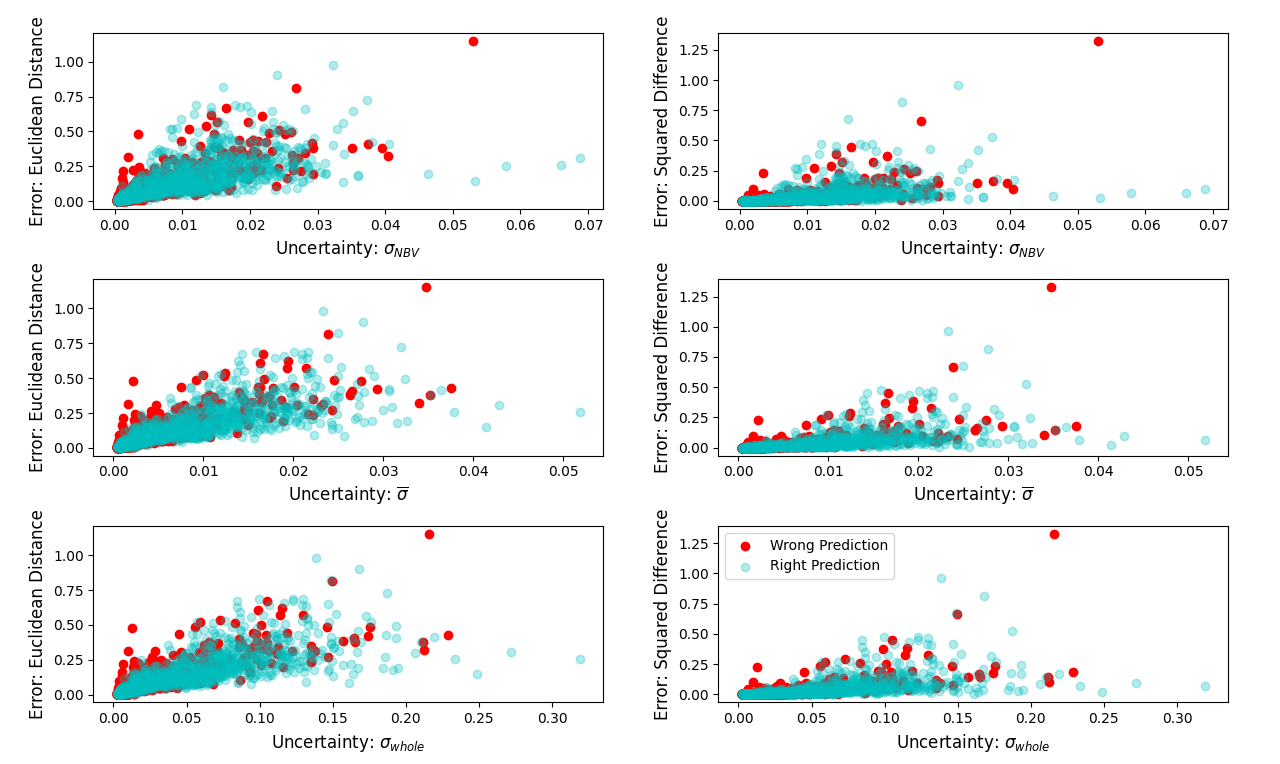}
    \caption{Distribution of the sample's error in function of uncertainty - different combinations. From top to bottom - uncertainty: $\sigma_{NBV}$, \eqref{eq:std_nbv}; $\overline{\sigma}$, \eqref{eq:std_mean}; $\sigma_{whole}$, \eqref{eq:std_whole}. From left to right - error: Euclidean Distance, \eqref{eq:dist_error}; Squared Differences, \eqref{eq:mse_error}.}
    \label{fig:error-modelling-combos}
\end{figure*}
\begin{table}[H]
\caption{Bayesian PC-NBV Model Loss, Model Error - Euclidean Distance and Squared Differences, and Model Accuracy}
\centering
\begin{tabular}{|c|c|c|c|c|}
\hline
\textbf{Bayesian}  & Loss, \eqref{eq:loss} &  Error, \eqref{eq:dist_error} & Error, \eqref{eq:mse_error} & Accuracy, \eqref{eq:model_accuracy} \\ \newline \textbf{PC-NBV}&&&&\\ 
\hline
Train                     & 0.287  & -                           & -                            & -        \\ \hline
Valid                     & 0.704  & 0.072                       & 0.018                        & 28.66\%  \\ \hline
Test                      & 0.662  & 0.071                       & 0.017                        & 29.02\%  \\ \hline
Test Novel                & 1.217  & 0.097                       & 0.035                        & 25.66\%  \\ \hline
\end{tabular}
\label{tab:bay-pc-nbv-results}
\end{table}

\subsection{Uncertainty Quantification}
\begin{figure}[H]
    \centering
    \includegraphics[width=\columnwidth]{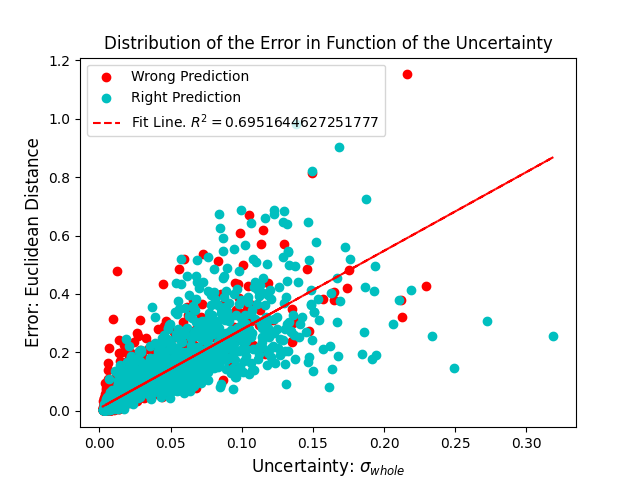}
    \caption{Distribution of the sample's Euclidean Distance error in function of the $\sigma_{whole}$ uncertainty and its approximated fit line (in red) with $R^2=0.695$.}
    \label{fig:error-unc-fit}
\end{figure}

% \begin{figure*}[h!]
% \captionsetup[subfloat]{labelfont=scriptsize,textfont=scriptsize,captionskip=0pt}
%   \subfloat[Distribution of right/wrong NBV predictions per uncertainty $\sigma_{accuracy}$ values \eqref{eq:std_accuracy}.]{
% 	\begin{minipage}[c][1\width]{
% 	   0.32\textwidth}
% 	   \centering
% 	   \includegraphics[width=1\textwidth]{Images/new_right_wrong_acc_unc_distrib.png}
%     \label{fig:acc-unc-dist}
% 	\end{minipage}}
%  \hskip 0pt plus .15fill	
%   \subfloat[Model Error using Euclidean Distance]{
% 	\begin{minipage}[c][1\width]{
% 	   0.32\textwidth}
% 	   \centering
% 	   \includegraphics[width=1\textwidth]{Images/new_err_unc_err.png}
%     \label{fig:error-unc-err}
% 	\end{minipage}}
%  \hskip 0pt plus .15fill	
%   \subfloat[Model accuracy \eqref{eq:model_accuracy} in function of uncertainty.]{
% 	\begin{minipage}[c][1\width]{
% 	   0.32\textwidth}
% 	   \centering
% 	   \includegraphics[width=1\textwidth]{Images/new_acc_unc_acc.png}
%     \label{fig:accuracy-unc-acc}
% 	\end{minipage}}
% \caption{On the middle, the Euclidean Distance, \eqref{eq:dist_error} in function of uncertainty $\sigma_{whole}$ \eqref{eq:std_whole}. On the right-hand side, uncertainty corresponds to $\sigma_{accuracy}$ \eqref{eq:std_accuracy}.}
% \end{figure*}

%The biggest challenge of uncertainty quantification is to find a measurement that correctly reflects the metric we want to predict.
We want to model the sample's error and the sample's accuracy, i.e - a large uncertainty value would mean that the prediction would have a large error and a higher probability of being wrong (of not having predicted the right NBV). The desired relation between both metrics would then be a directly proportional one. A purely horizontal or vertical pattern would not be good to differentiate between right and wrong predictions. To study this and understand which uncertainty measurement - \eqref{eq:std_mean}, \eqref{eq:std_nbv} or \eqref{eq:std_whole} - best models which error metric - \eqref{eq:dist_error} or \eqref{eq:mse_error} - a plot of the error by the uncertainty of each sample for the different combinations was made, and is displayed in Figure \ref{fig:error-modelling-combos}. In addition, we want to study if the accuracy of the selected view can be modeled as a function of the uncertainty. To do that, we plot each sample with one of two colors: blue if presents a right NBV prediction and red if is wrong.

Figure \ref{fig:error-modelling-combos} shows that the approximation of the desired proportional relation occurs when the error metric is given by Euclidean Distances. The uncertainty metric that allows a better distribution and has a larger range of values is $\sigma_{whole}$.

%When the metric is given by the Squared Differences the distribution is quite horizontal, with no clear distinction in error as uncertainty increases. Hence, the chosen error metric to model is going to be the Euclidean Distance \eqref{eq:dist_error}.

%Looking at the left side of the Figure \ref{fig:error-modelling-combos}, where the used error metric is the Euclidean Distance \eqref{eq:dist_error}, it is possible to verify that the uncertainty metrics that better allow the desired proportional distribution are $\overline{\sigma}$ \eqref{eq:std_mean} and $\sigma_{whole}$ \eqref{eq:std_whole}. Both of these metrics allow quite similar distributions. However, $\sigma_{whole}$ has a larger range of uncertainty values. While the values of $\overline{\sigma}$ only vary in the interval $[0, 0.06]$ the $\sigma_{whole}$ values vary in the interval $[0, 0.4]$ making it a more suitable metric.

Figure \ref{fig:error-unc-fit} shows a closer look at the distribution of the error \eqref{eq:dist_error} in function of the uncertainty $\sigma_{whole}$ - \eqref{eq:std_whole}, as well as the line that best fits the data. This fit is evidently not perfect, having a $R^2 = 0.695$. However, it serves to highlight the directly proportional tendency of the data distribution.

The graphs show that $\sigma_{whole}$ \eqref{eq:std_whole} is not related to the accuracy of the sample \eqref{eq:sample_accuracy}. The desired behavior would be for the blue dots (the right predictions) to be distributed in the lower range of the uncertainty values and the red dots (the wrong predictions) in the higher range - which is not visible.

%This happens because the accuracy of a sample \eqref{eq:sample_accuracy} is not correlated to its error \eqref{eq:dist_error}. The accuracy measures if the highest coverage score was attributed to the groundtruth's NBV while the error measures how close the coverage scores are from the groundtruth's coverage scores. As the $\sigma_{whole}$ uncertainty calculation is based on the MC samples $Euclidean\:Distance$ error, it is natural that it can correctly offer an insight of the error, not being able to do so for the sample's accuracy.

An uncertainty metric capable of modeling the sample's accuracy \eqref{eq:sample_accuracy} can be obtained if it is calculated based on the the MC samples accuracy.  
We computed the accuracy of every MC sample for uncertainty estimation as
\begin{equation}
    accuracy(mc_i) = \begin{cases}
        1, NBV_{mc_{i}} = NBV_{fp} \\
        0, otherwise
    \end{cases}
    \label{eq:samples_accuracy}
\end{equation}
\begin{equation*}
    i \in \{1,...40\}
\end{equation*}
This accuracy checks if the sample's NBV is the same NBV of the final prediction \eqref{eq:final_pred}. The uncertainty is calculated as the Standard Deviation of the MC sample's accuracy:
\begin{equation}
    \sigma_{accuracy} = \sqrt{\frac{1}{n_{mc}}\sum_{i=1}^{n_{mc}} (1-accuracy(mc_{i}))^2}
    \label{eq:std_accuracy}
\end{equation}

\begin{figure}[H]
    \centering
    \includegraphics[width=\columnwidth]{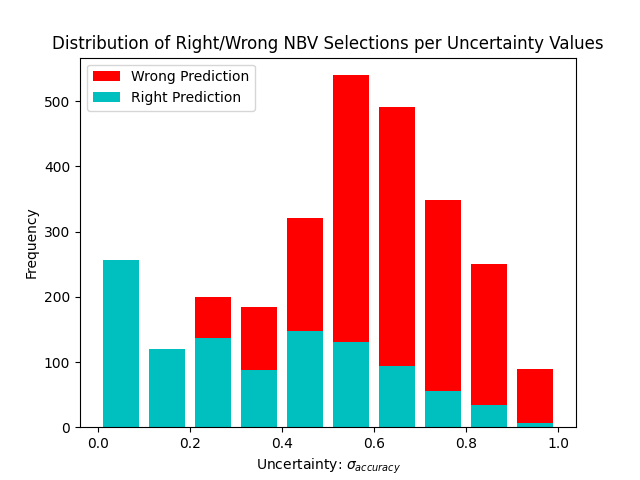}
    \caption{Distribution of right/wrong NBV predictions per uncertainty $\sigma_{accuracy}$ values \eqref{eq:std_accuracy}.}
    \label{fig:acc-unc-dist}
\end{figure}

Figure \ref{fig:acc-unc-dist} shows an histogram that quantifies how many samples in each uncertainty bin get the NBV prediction right and how many get it wrong. Most of the samples in the low range of the $\sigma_{accuracy}$ uncertainty values are samples that got the prediction right, while most of the samples with higher uncertainty values are samples that weren't able to do so. This was the desired behavior for accuracy modeling.
%and validates the proposition that in order for uncertainty to reflect accuracy it needed to be computed from values that measured this quantity.

By now we have two uncertainty metrics capable of estimating uncertainty measurements that give insights into the prediction's error - $\sigma_{whole}$ - and into the prediction's accuracy - $\sigma_{accuracy}$. However, an important question arises: \textbf{How can these metrics help improve the model's performance?}
\begin{table*}[t!]
\centering
\caption{New Coverage Score Formulas - Best Results}
\label{tab:formulas_results}
\begin{tabular}{cc|c|c|c|c|}
\cline{3-6}
 &  & Error \eqref{eq:dist_error} & Error Change & Accuracy (\%) \eqref{eq:model_accuracy} & Accuracy Change \\ \hline
\multicolumn{1}{|c|}{{$fp \pm \left( \sigma_{view} \times (\sigma_{normalized} \times 0.8) \right)$}} & Valid & \begin{tabular}[c]{@{}c@{}}0.073\\ 0.072\end{tabular} & \begin{tabular}[c]{@{}c@{}}+0.001\\ 0\end{tabular} & \begin{tabular}[c]{@{}c@{}}28.45\\ 28.71\end{tabular} & \begin{tabular}[c]{@{}c@{}}-0.22\\ +0.05\end{tabular} \\ \cline{2-6} 
\multicolumn{1}{|c|}{} & Test & \begin{tabular}[c]{@{}c@{}}0.072\\ 0.072\end{tabular} & \begin{tabular}[c]{@{}c@{}}0\\ 0\end{tabular} & \begin{tabular}[c]{@{}c@{}}29.16\\ 29.21\end{tabular} & \begin{tabular}[c]{@{}c@{}}+0.14\\ +0.19\end{tabular} \\ \cline{2-6} 
\multicolumn{1}{|c|}{} & Test Novel & \begin{tabular}[c]{@{}c@{}}0.097\\ 0.097\end{tabular} & \begin{tabular}[c]{@{}c@{}}0\\ 0\end{tabular} & \begin{tabular}[c]{@{}c@{}}25.80\\ 25.64\end{tabular} & \begin{tabular}[c]{@{}c@{}}+0.14\\ -0.02\end{tabular} \\ \hline
\multicolumn{1}{|c|}{{$fp \pm \left( \sigma_{view} \times \sigma_{normalized} \times \sigma_{accuracy}^{0.4} \right)$}} & Valid & \begin{tabular}[c]{@{}c@{}}0.073\\ 0.072\end{tabular} & \begin{tabular}[c]{@{}c@{}}+0.001\\ 0\end{tabular} & \begin{tabular}[c]{@{}c@{}}28.45\\ 28.74\end{tabular} & \begin{tabular}[c]{@{}c@{}}-0.21\\ +0.08\end{tabular} \\ \cline{2-6} 
\multicolumn{1}{|c|}{} & Test & \begin{tabular}[c]{@{}c@{}}0.072\\ 0.072\end{tabular} & \begin{tabular}[c]{@{}c@{}}0\\ 0\end{tabular} & \begin{tabular}[c]{@{}c@{}}29.21\\ 29.18\end{tabular} & \begin{tabular}[c]{@{}c@{}}+0.19\\ +0.16\end{tabular} \\ \cline{2-6} 
\multicolumn{1}{|c|}{} & Test Novel & \begin{tabular}[c]{@{}c@{}}0.096\\ 0.097\end{tabular} & \begin{tabular}[c]{@{}c@{}}-0.001\\ 0\end{tabular} & \begin{tabular}[c]{@{}c@{}}25.77\\ 25.66\end{tabular} & \begin{tabular}[c]{@{}c@{}}+0.11\\ +0.00\end{tabular} \\ \hline
\end{tabular}
\end{table*}

\subsection{Uncertainty for Model Performance Improvement}

\subsubsection{Coverage Score Improvement}
It is possible to improve performance incorporating the uncertainty predictions with the coverage scores predictions, in an attempt of generating an improved coverage score. This coverage score would ideally be more similar to the groundtruth, reducing the Euclidean Distance error \eqref{eq:dist_error}. It could also improve the coverage score of the real NBV, increasing that view's coverage score, therefore improving the sample's accuracy \eqref{eq:sample_accuracy}.
%Several different combinations incorporating the uncertainty measurements with the coverage prediction were evaluated. 
The formulas that presented better results, being able to slightly increment the model's accuracy, are presented in Table \ref{tab:formulas_results}. $\sigma_{normalized}$ is given by $\sigma_{view} / max(\sigma_{view})$. $max(\sigma_{view})$ is the maximum measured $\sigma_{view}$ throughout all the test samples, for each view. To get to these values we need to have prior information of all the view's uncertainty. To achieve this a calibration set can be used to obtain an estimate of the $max(\sigma_{view})$ factor.

\begin{figure}[H]
    \centering
    \includegraphics[width=\columnwidth]{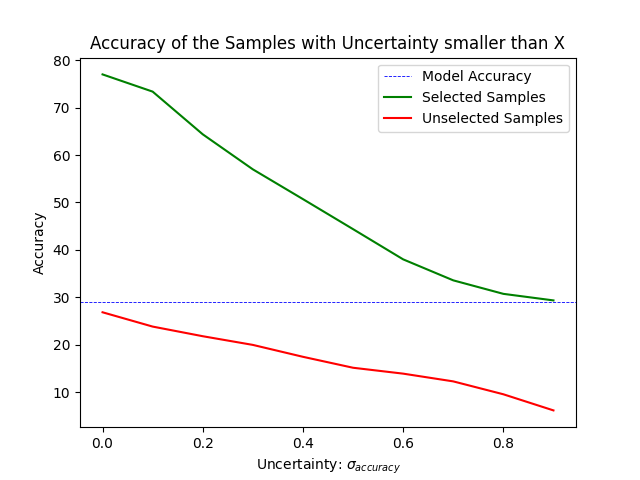}
    \caption{Model accuracy \eqref{eq:model_accuracy} in function of uncertainty $\sigma_{accuracy}$ \eqref{eq:std_accuracy}.}
    \label{fig:accuracy-unc-acc}
\end{figure}
\subsubsection{Decision Support}
\begin{figure}[H]
    \centering
    \includegraphics[width=\columnwidth]{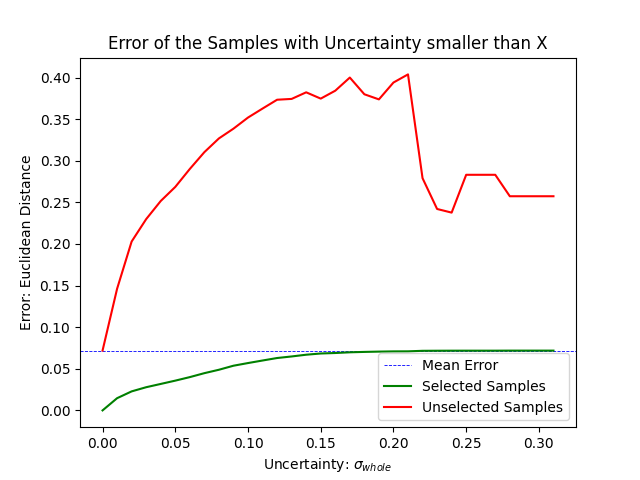}
    \caption{Model Error (Euclidean Distance, \eqref{eq:dist_error} in function of uncertainty $\sigma_{whole}$ \eqref{eq:std_whole}.}
    \label{fig:error-unc-err}
\end{figure}

By rejecting predictions with an uncertainty value higher than a threshold, we will discard low quality predictions while ensuring a better performance.
Figure \ref{fig:error-unc-err} shows, in green, the plot of the model Euclidean Distance error when considering only samples with $\sigma_{whole}$ uncertainty smaller or equal than the value in the x axis. In red it plots the model error considering all the other samples that were not used, i.e. that had a higher uncertainty value. The dotted blue line marks the model's error when considering all the samples.
This graph shows that in fact when we only account the samples with low uncertainty the model achieves lower error values. If we consider samples with uncertainty lower than 0.15, it is possible to reduce the model's error.

Figure \ref{fig:accuracy-unc-acc} shows the same behavior but for the model's accuracy. It becomes possible to achieve extremely good accuracy percentages comparing to the original model accuracy simply by discarding predictions with high uncertainty. In this case if we want to achieve a model accuracy of 60\% we need to discard the samples with an $\sigma_{accuracy}$ uncertainty higher than 0.2.

To use this approach there is a need to set an uncertainty limit that will classify a prediction as acceptable or not. This limit can vary depending on the desired model performance and on the used data. The limit should be calibrated on a calibration set, where one would examine which uncertainty values allowed the desired model performance.

To use this method it would be, however, necessary to implement what to do when a prediction is discarded during the 3D acquisition process, since the model would not provide any information of what the NBV would be. Then tests to show the efficiency of the implemented method in comparison to the efficiency of simply choosing a wrong predicted NBV needed to be run. This is an essential analysis to evaluate the usage relevance of these measurements as decision support metrics.
An example of what could be done to avoid not having a NBV to go to when a prediction is discarded is to choose the viewpoint that least interferes with the prediction's uncertainty.

\section{Conclusion and Future Work}

% The Monte Carlo Dropout method was successfully applied to the PC-NBV model, allowing the model to estimate its own prediction uncertainty. By using the uncertainty measurements as support-decision metrics we were able to significantly improve the model's accuracy from 30\% to 60\%-80\%. However, this approach requires further investigation into how to handle cases where the model discards a prediction due to high uncertainty.

% Future works should explore the effect of different dropout layer configurations, and of different dropout probabilities and number of Monte Carlo samples. And additionally test the use of the model in a real-world scenario.

% ----VERSÃO MAIS COMPLETA ----------
The Monte Carlo Dropout method was successfully implemented in the PC-NBV model. By applying dropout layers after all the convolutional layers of the model and sampling 40 model predictions for the same input during inference, a functional framework to obtain the prediction's uncertainty was implemented. Several uncertainty metrics were studied with the goal of finding the ones that better reflected the prediction's error and accuracy. It was not possible to find a metric that combined the two, but two separate metrics that successfully reflected these quantities were found - $\sigma_{whole}$ and $\sigma_{accuracy}$ 

Moreover, the usage potential of the uncertainty measurements to leverage the model's performance, was studied.
By implementing a new coverage score formula that incorporates the uncertainty measurements with the final prediction we were able to obtain some slight model error and accuracy improvements. On the other hand, by discarding any sample that presented an high uncertainty value we were able to improve the model's error, but most important, the model's accuracy. With this approach it was possible to improve the model's accuracy from 30\% to 60\%-80\%. However, this method faces an obstacle that needs further studies, as it would need another NBV decision approach to use when the model discards a prediction.

In order to improve the current work, in the future it would be interesting to carry some additional analysis such as the study of the effect different dropout layer placements could achieve, as well as different dropout probabilities and number of Monte Carlo samples. Lastly another important test to carry would be the use of this model on a real life scenario.
%-----------------------------

%The reviewed works \cite{bayesian_pose-net} and \cite{bayesian_seg-net} showed that in fact applying dropout after each convolutional layer could work as a too strong of a regularizer, and stated that only applying dropout layers in certain parts of their networks showed better results. That analysis was not done in this work, but could also be an interesting study.
%As well as the optimization of the dropout probability, that in this work was left at 0.5. However, it is possible that another probability could achieve better results. Also, another study that was not covered was the study of the ideal number of Monte Carlo dropout samples, which is a parameter that could also be optimized. Lastly, another important study to run in the future is the use of this model on a real life scenario.

\vfill

\end{document}